\documentclass{article}

\usepackage{arxiv}
\usepackage{float}
\usepackage[utf8]{inputenc} 
\usepackage[T1]{fontenc}    
\usepackage{hyperref}       
\usepackage{url}            
\usepackage{booktabs}       
\usepackage{amsmath}
\usepackage{amsfonts}       
\usepackage{nicefrac}       
\usepackage{microtype}      
\usepackage{cleveref}       
\usepackage{lipsum}         
\usepackage{graphicx}
\usepackage[numbers,square]{natbib}
\usepackage{doi}

\title{A Comparison of Traditional Machine Learning Algorithms and LSTM-Based Deep Learning Models for Email Sentiment Analysis}

\date{}

\newif\ifuniqueAffiliation
\uniqueAffiliationtrue

\ifuniqueAffiliation 
\author{ {\hspace{1mm}Virdio Samuel Saragih}\\
	Faculty of Science\\
	Sumatra Institute of Technology\\
	\texttt{Virdio.122450124@student.itera.ac.id} \\
	\And
	{\hspace{1mm}Baruna Abirawa} \\
	Faculty of Science\\
	Sumatra Institute of Technology\\
    \texttt{baruna.122450097@student.itera.ac.id} \\
    \And
	{\hspace{1mm}Kartini Lovian Simbolon} \\
    Faculty of Science\\
	Sumatra Institute of Technology\\
    \texttt{kartini.122450097@student.itera.ac.id} \\
    \And
	{\hspace{1mm}Luluk Muthoharoh} \\
    Faculty of Science\\
	Sumatra Institute of Technology\\
    \texttt{luluk.muthoharoh@sd.itera.ac.id} \\
     \And
	{\hspace{1mm}Ardika Satria} \\
    Faculty of Science\\
	Sumatra Institute of Technology\\
    \texttt{ardika.satria@sd.itera.ac.id} \\
    \And
	{\hspace{1mm}Martin C.T. Manullang} \\
    Faculty of Science\\
	Sumatra Institute of Technology\\
	    \texttt{martin.manullang@if.itera.ac.id} \\    
}

\fi

\renewcommand{\headeright}{}
\renewcommand{\undertitle}{}
\renewcommand{\shorttitle}{}

\begin{document}
\maketitle

\begin{abstract}
The rapid growth of electronic communication has necessitated more robust systems for email classification and sentiment detection. This study presents a comparative performance analysis between traditional machine learning algorithms and deep learning architectures, specifically focusing on Support Vector Machines (SVMs), Logistic Regression, Naive Bayes, and Long Short-Term Memory (LSTM). Utilizing Word2Vec embeddings for feature representation, our experimental results indicate that the SVM model with a linear kernel achieves the highest efficiency and accuracy, reaching a peak performance of 98.74\%. While the LSTM model demonstrates exceptional recall capabilities in detecting spam-related sentiments, it requires significantly more computational time compared to discriminative statistical models. Detailed evaluations via confusion matrices further reveal that traditional classifiers remain highly robust for dense vector spaces. This research concludes that  email detection tasks, SVM offers the most optimal balance between predictive precision and processing speed. These findings provide critical insights for developing high-performance automated email filtering systems in professional and academic environments.
\end{abstract}

\keywords{Email Sentiment Analysis \and Email Detection \and Machine Learning \and Deep Learning \and SVM \and LSTM \and Word2Vec}

\section{Introduction}
In this era of rapidly advancing technology,
email remains one of the most widely used
communication tools, particularly in professional settings \cite{kautsar2025analisis}. This is due to email’s simplicity
and speed; however, the increasing volume of email
has led to negative consequences by creating opportunities
for scammers to misuse email by
spreading spam containing phishing attempts or malware \cite{manguma2024analisis}. 
Which can result in data and information leaks. Such spam
is sent repeatedly, there by disrupting and
harming email account holders \cite{adduali2026komparasi}.

Therefore, email processing is required to classify
emails into spam and non-spam by utilizing
Natural Language Processing and employing
machine learning and deep learning approaches. NLP is a technique
for transforming text that is initially incomprehensible to
computers into a processable format through preprocessing
and feature extraction, where text is represented as
numerical values \cite{pais2022nlp}. Whereas machine learning falls under
supervised learning, where models learn by recognizing
patterns in labeled data \cite{isnansyah2024analisis}. On the other hand,
deep learning methods such as LSTM are widely
used for classification because the algorithm has
proven effective at retaining both the longest
and shortest information, as it is derived from
the RNN algorithm \cite{wicaksana2026peningkatan}.

Some previous researchers have used NLP and
combined it with machine learning and deep learning models
such as \cite{isnansyah2024analisis}. Finding that the Naive Bayes algorithm
delivered the best performance with an accuracy of 93.3\%,
precision of 90.91\%, recall of 96.77\%, and an F1-score of 93.75%
compared to the Support Vector Machine
and LSTM, while another study \cite{ainun2024klasifikasi}. Found
the SVM model to be the best with 99\% accuracy \cite{reviantika2021analisis}. And
another researcher achieved 95\% accuracy in
classifying SMS into spam and non-spam using
logistic regression; meanwhile, another researcher using
deep learning, such as the researcher in \cite{wicaksana2026peningkatan}. Achieved
optimized accuracy of 95.65\%, precision of 93.65\%,
recall of 98.33\%, and an F1-Score of 95.93\% by using
the LSTM method with word2vec feature extraction and
parameter optimization via grid search.

Based on the aforementioned prior research, this study will perform sentiment analysis on an email dataset to classify emails into two categories spam and non spam by testing machine learning and deep learning algorithms. For machine learning, three methods will be used: Naive Bayes, SVM, and logistic regression; while for deep learning, LSTM will be employed. All four of these algorithms utilize Word2vec feature extraction. The objective of this study is to evaluate the performance of these algorithms in classifying emails and to determine which algorithm delivers the best performance.

\section{Related Work}
\label{sec:headings}

\subsection{Email}
Email (electronic mail) is a form of digital communication used to send and receive messages via computer networks or the internet. In today’s world, email serves not only as a tool for exchanging information, but also as the primary means of communication in a wide range of personal, business, and academic activities. Email is an effective and efficient method of communication for sharing data and information electronically, offering easy access and a high capacity for message replication \cite{baroto2021email}.

\subsection{Sentiment Analysis}
Sentiment analysis is commonly used to extract information and analyze public opinion both positive and negative contained within text. This approach involves natural language processing and data mining so that the system can understand user sentiment and classify it into specific categories. Additionally, sentiment analysis can also be applied to evaluate perceptions regarding a specific topic, rather than just individual opinions, thereby capturing the evolving dynamics of a discussion \cite{hakim2021analisa}\cite{nakov2016semeval}. Sentiment analysis is a method used to identify and describe opinions within a text whether objective or subjective that exhibit a particular sentiment. This method is based on text analysis techniques to determine the level of subjectivity in an opinion, for example, in data from social media platforms such as Twitter. Thus, sentiment analysis can be used to understand users’ emotions or attitudes and draw conclusions based on the linguistic patterns used in a post \cite{hakim2021analisa}\cite{rajputnlpsentimentclinical}.

\subsection{Support Vector Machine}
Support Vector Machines  are a method in machine learning used for prediction, whether for classification or regression problems. However, to handle data that cannot be linearly separated, this method has been further developed through the use of kernel functions, which allow data to be transformed into a higher dimensional space, thereby making class separation more feasible \cite{hendriyana2022analisis}. This approach works by determining an optimal decision boundary in the form of a hyperplane, which is capable of maximizing the margin between two different classes \cite{muttaqin2021analisis}.

\subsection{Regretion logistics}
Logistic regression is a machine learning approach used to solve classification problems with categorical outputs. This method works by modeling the probability of a data point belonging to a specific class, thereby enabling the determination of whether a data point falls into a particular category, such as positive or negative sentiment \cite{reviantika2021analisis}. In the context of this study, the use of logistic regression is considered appropriate due to its ability to efficiently handle high-dimensional data, such as text. Furthermore, this algorithm is highly effective for binary classification cases, where there are only two possible classes, making it relevant to the needs of the analysis being conducted\cite{elmaliyasari2025deteksi}.

\subsection{Naive Bayes}
Naive Bayes is a classification method in machine learning based on a probabilistic approach. This algorithm is widely used in sentiment analysis due to its ability to process high-dimensional text data in a relatively simple yet effective manner. The main principle of this method is to estimate the probability that a document or text belongs to a specific category, such as positive or negative sentiment, based on the probability distribution of the words that compose it. In its calculations, Naive Bayes assumes that each feature or word is independent of the others; this simplification allows for faster computation without significantly sacrificing performance \cite{elmaliyasari2025deteksi}.

\subsection{LSTM}
LSTM ability to retain historical information makes it particularly well-suited for handling sequential data, including text, as the model can learn long-term dependencies between words or sentences. As a result, LSTM is widely used in various research areas related to natural language processing, particularly for classification tasks and sentiment analysis. The LSTM architecture consists of units known as memory cells, which are equipped with several control components: the input gate, the forget gate, and the output gate. These three components play a role in regulating the flow of information entering, stored, and output from the system, enabling the model to select relevant information and disregard irrelevant details. This approach effectively addresses the vanishing gradient problem that frequently occurs in traditional RNNs \cite{alghifari2022bilstm}.

\section{Dataset}
The dataset used in this study is the “Dataset of Spam Email in Indonesian Language,” obtained from the open-source platform Kaggle. This dataset has been translated into Indonesian and is available in Comma-Separated Values (CSV) format. Structurally, the dataset consists of two main attributes: the “message” column, which contains the email content, and the “label” column, which indicates the category of each message. 

\begin{table}[h]
\centering
\caption{Email Spam Classification}
\begin{tabular}{lcl}
\toprule
\textbf{Kategori} & \textbf{Pesan} \\
\midrule
Spam & Apakah Anda kalah? Jawabannya akan membuat Anda takjub!... \\
Spam & bro lihat produk baru yang luar biasa ini, semoga Anda bisa lebih baik?... \\
Spam & Dapatkan tagihan popok bayi Anda dibayar selama setahun... \\
Ham & Musim Panas di Enron Hi Vince: Jika Anda atau departemen sumber daya manusia... \\
Ham & Persetujuan Peninjau Harap dicatat bahwa karyawan Anda telah menyarankan... \\
\bottomrule
\end{tabular}
\end{table}

The label is categorical with two classes: spam and non-spam. The total data used in this study consists of 2,620 entries, comprising 1,363 messages labeled as spam and 1,258 messages labeled as non-spam. This composition indicates that the dataset has a relatively balanced class distribution, making it sufficiently representative for use in the training and evaluation of classification models.

\section{Methodology}

\subsection{Preprocessing}
In this stage, the text data is processed to remove unnecessary elements and standardize the formatting. This is necessary to make the data cleaner, more consistent, and ready for use in the modeling stage \cite{husada2021analisis}. Before the email texts are fed into the Machine Learning and Deep Learning models, they must undergo a preprocessing phase. The objective is to clean the data from noise and standardize the text format so the model can extract features optimally. In this study, the text preprocessing pipeline consists of three sequential stages:

\subsubsection{Text Cleaning}
The first stage aims to eliminate irrelevant text elements such as punctuation, links, and email-specific noise. The implemented steps include:
\begin{enumerate}
    \item Case Folding: Converting all characters in the text to lowercase.
    \item URL removal: Eliminating all uniform resource locators (URLs) and hyperlinks (e.g., http:// or https://).
    \item Email Attribute Removal: Removing email adressses, domain extensions (e.g., .com, .net, .id), and default email client subject prefixes (e.g., re:, fw:, fwd:).
    \item Character Filtering: Stripping all characters other than Indonesian alphabetic letters. Numbers and punctuation marks are entirely removed.
    \item Whitespace Trimming: Removing double or excessive whitespace, leaving only a single space between words.
\end{enumerate}




\subsubsection{Stopword Removal}
Stopwords are common conjunctions or frequently used words that do not carry significant semantic meaning to distinguish between spam and non-spam classes (e.g., "yang" (which), "dan" (and), "di" (in/at)). In this system, stopword removal is implemented using the Sastrawi library (StopWordRemoverFactory), which is specifically designed for the Indonesian language. The cleaned and normalized text is split into tokens (words), and any token present in the Sastrawi stopword list is discarded. The final output of these three stages is a sequence of clean, standardized, and meaningful core tokens, ready for the subsequent feature extraction process (e.g., using Word2Vec).

\subsection{Feature Extraction}
\subsubsection{Machine learning}
Cleaned text are tokenized and each token is embedded with Word2Vec into a 100-D vector so related words tend to be nearby in space instead of being represented only by independence assumptions like raw counts. The model may be loaded or trained on the training split; token vectors for one message are averaged into one document vector, which classical learners (logistic regression, naive Bayes, linear SVM) take as input.

\subsubsection{Deep learning}
Tokens map to integer ids from a training-only vocabulary with a frequency cutoff, sequences are padded or truncated to fixed length. A trainable embedding maps ids to dense vectors that the LSTM and classifier learn jointly, so the main "features" are indexes plus alignment to fixed length, the useful representation is learned inside the network from the final sequence state.

\subsection{Model Architectures}
This study implements and compares both classical Machine Learning models and a Deep Learning approach for the classification task

\subsubsection{Machine Learning Models}
The system evaluates three baseline traditional algorithms: Support Vector Machine (SVM), Naive Bayes, and Logistic Regression. These models are trained using the extracted Word2Vec features to establish a performance benchmark.

\subsubsection{Deep Learning Models}
To capture long-term dependencies and sequence patterns in the text, a Long Short-Term Memory (LSTM) neural network architecture is implemented (LSTMSpamModel). The network consists of the following stacked layers:

\begin{enumerate}
    \item \textbf{Input preprocessing:} Each email is tokenized, mapped to vocabulary indices (with \texttt{\textless UNK\textgreater} for unknown tokens), and padded or truncated to a fixed sequence length of 50, yielding input tensors of shape (\textit{batch\_size}, 50) integer token IDs.
    
    \item \textbf{Embedding layer:} A trainable \texttt{nn.Embedding} maps each token ID to a dense vector of dimension 64 
    (\textit{embed\_dim}=64), with padding index 0 ignored for the embedding of padded positions. The resulting sequence tensor has shape (\textit{batch\_size}, 50, 64).
    
    \item \textbf{LSTM layer:} A single bidirectional-off LSTM (\textit{nn.LSTM}) with input size 64 and hidden size 64 (\textit{hidden\_dim}=64, \textit{batch\_first=True}), processes the embedded sequence and produces hidden states for every timestep.
    \item \textbf{Sequence pooling:} Only the hidden state from the \emph{last} LSTM layer at the final timestep is used as the sentence representation (corresponding to \textit{hidden[-1]} in the implementation).
    \item \textbf{Output layer:} A linear layer maps the 64-dimensional hidden state to a single logit, followed by a Sigmoid activation so the model outputs a probability in \((0,1)\) for the spam class (\textit{BCELoss} is used during training).
\end{enumerate}

\subsubsection{Training and Hyperparameter}
The dataset is split into training and testing sets (typically using an 80:20 ratio). For the LSTM model training, the Binary Cross-Entropy Loss (BCEWithLogitsLoss / BCELoss) is utilized to calculate the error since this is a binary classification problem. The model is optimized using the Adam Optimizer.

\begin{table}[h]
\centering
\caption{LSTM model hyperparameters}
\label{tab:lstm-hparams}
\begin{tabular}{ll}
\toprule
\textbf{Hyperparameter} & \textbf{Value} \\
\midrule
Embedding dimension (\(d_{\mathrm{emb}}\)) & 64 \\
LSTM hidden dimension (\(d_{\mathrm{hid}}\)) & 64 \\
LSTM layers & 1 \\
\texttt{batch\_first} & True \\
Padding token index (\texttt{padding\_idx}) & 0 \\
Unknown token index (\texttt{\textless{}UNK\textgreater{}}) & 1 \\
Max sequence length (\(T_{\max}\)) & 50 \\
Vocabulary min frequency (train) & 2 \\
Vocabulary size (\(\lvert \mathcal{V} \rvert\)) & \(\lvert \mathcal{V} \rvert = \texttt{len(vocab)}\) \\
Batch size & 32 \\
Optimizer & Adam \\
Learning rate & \(1 \times 10^{-3}\) \\
Loss & BCELoss \\
Epochs & 30 \\
Device & CUDA (if available), else CPU \\
\bottomrule
\end{tabular}
\end{table}

Throughout the training loop, metrics such as training loss, validation accuracy, and F1-score are continuously monitored and logged using Weights \& Biases (WandB) for comprehensive experiment tracking and hyperparameter tuning visualization. The trained model weights are subsequently saved as a PyTorch artifact (spam\_model\_lstm.pth).

\subsubsection{Evaluation Metrics}
The performance of the trained models is evaluated on the unseen test set using the following standard classification metrics:

\begin{enumerate}
    \item Accuracy: The ratio of correctly predicted observation (both true spam and true ham) to the total observations.
        \begin{equation}
        \text{Accuracy} = \frac{\text{TP} + \text{TN}}{\text{TP} + \text{TN} + \text{FP} + \text{FN}}
        \end{equation}
    \item Precision: The ratio of correctly predicted positive observation to the total predicted positives (crucial for minimizing false positive/legitimate emails marked as spam).
        \begin{equation}
        \text{Precision} = \frac{\text{TP}}{\text{TP} + \text{FP}}
        \end{equation}
    \item Recall: The ratio of correctly predicted positive observation to all actual positives in the actual class.
        \begin{equation}
        \text{Recall} = \frac{\text{TP}}{\text{TP} + \text{FN}}
        \end{equation}
    \item F1-Score: The weighted average of Precision and Recall. It is especially useful when the class distribution is uneven
        \begin{equation}
        \text{F1-Score} = 2 \times \frac{\text{Precision} \times \text{Recall}}{\text{Precision} + \text{Recall}}
        \end{equation}
\end{enumerate}


\section{Experiments}

\begin{table}[h]
\centering
\caption{Dataset split for the experiment.}
\label{tab:dataset-split}
\begin{tabular}{lcr}
\toprule
\textbf{Subset} & \textbf{Number of Samples} & \textbf{Percentage} \\
\midrule
Training Data & 2,068 & 80\% \\
Test Data & 517 & 20\% \\
\textbf{Total Clean Data} & \textbf{2,585} & \textbf{100}\textbf{\%} \\
\bottomrule
\end{tabular}
\end{table}

All experiments use the cleaned Indonesian email corpus restricted to non-empty messages and two classes, ham and spam. The data partition in Table~\ref{tab:dataset-split} is a stratified random split into 80\% training and 20\% testing so that class proportions are preserved; a fixed random seed ensures repeatable splits.

\subsection{Classical machine-learning baselines}
The classical setup follows a conventional supervised workflow. Messages undergo the same text normalization and filtering as above. Each document is represented by dense Word2Vec features: word vectors are aggregated into a single document vector by averaging over tokens. Three discriminative models are trained on these representations: logistic regression with a sufficiently large iteration limit for convergence, Gaussian naive Bayes, and a linear support vector machine.

\subsection{Deep learning baseline}
The neural baseline is a compact recurrent model in PyTorch. Training text is used to build a subword-level vocabulary with a minimum token frequency threshold; rare tokens map to a single unknown index and padding uses a dedicated index. Each message is tokenized into lowercase alphabetic tokens, converted to index sequences, and padded or truncated to a fixed maximum length of 50. Mini-batches of size 32 feed a trainable embedding of dimension 64, followed by one LSTM layer with 64 hidden units; the last hidden state is passed through a linear layer and a sigmoid to produce a spam probability. Optimization uses Adam with learning rate 0.001 and binary cross-entropy loss for 30 epochs. Only the average training loss per epoch is monitored during optimization; post-hoc test metrics such as accuracy or F--measure are not reported automatically in this configuration, though a held-out split is reserved in the same proportion as in Table~\ref{tab:dataset-split}.

\begin{table}[h]
\centering
\caption{Machine learning \& LSTM model configuration.}
\label{tab:model-config}
\begin{tabular}{l c c p{5cm}}
\toprule
\textbf{Model} & \textbf{Configuration} \\
\midrule
SVM & LinearSVC \\
Logistic Regression & LogisticRegression(max\_iter=1000) \\
Naive Bayes & GaussianNB \\
\bottomrule
\end{tabular}
\end{table}

\section{Results and Discussion}
\subsection{Machine Learning Model Results}
Three machine learning models (SVM - Linear Kernel, Logistic Regression, and Naive Bayes) are evaluated on the same Word2Vec features. Validation results are used for model selection, and test results for generalization assessment.

\begin{table}[h]
\centering
\caption{PyCaret model compariso}
\label{tab:model-comparison}
\resizebox{\linewidth}{!}{
\begin{tabular}{lccccccccc}
\toprule
\textbf{Model} & \textbf{Type} & \textbf{Accuracy} & \textbf{AUC} & \textbf{Recall} & \textbf{Precision} & \textbf{F1} & \textbf{Kappa} & \textbf{MCC} & \textbf{TT (Sec)} \\
\midrule
SVM & SVM - Linear Kernel & 0.9874 & 0.9990 & 0.9874 & 0.9876 & 0.9874 & 0.9747 & 0.9749 & 0.908 \\
LR  & Logistic Regression & 0.9753 & 0.9982 & 0.9753 & 0.9763 & 0.9753 & 0.9504 & 0.9514 & 2.695 \\
NB  & Naive Bayes         & 0.9449 & 0.9441 & 0.9449 & 0.9456 & 0.9448 & 0.8892 & 0.8901 & 1.988 \\
\bottomrule
\end{tabular}
}
\end{table}

SVM (Linear Kernel) is the best performing machine learning model, achieving an outstanding 98.74\% accuracy and a 98.74\% F1-score. It also trains exceptionally fast at 0.908 seconds, demonstrating that the SVM linear kernel handles Word2Vec embeddings highly effectively. Logistic Regression serves as a strong alternative, performing slightly below SVM with 97.53\% accuracy and a 97.53\% F1-score. However, it requires the longest training time among the three models at 2.695 seconds, making it less efficient than SVM. Naive Bayes performs the worst among the evaluated models, yielding 94.49\% accuracy and a 94.48\% F1-score. These results indicate that Naive Bayes is less suited for the dense, continuous vector representations generated by the Word2Vec features.

\begin{figure}[H]
  \centering
  \includegraphics[width=0.6\textwidth]{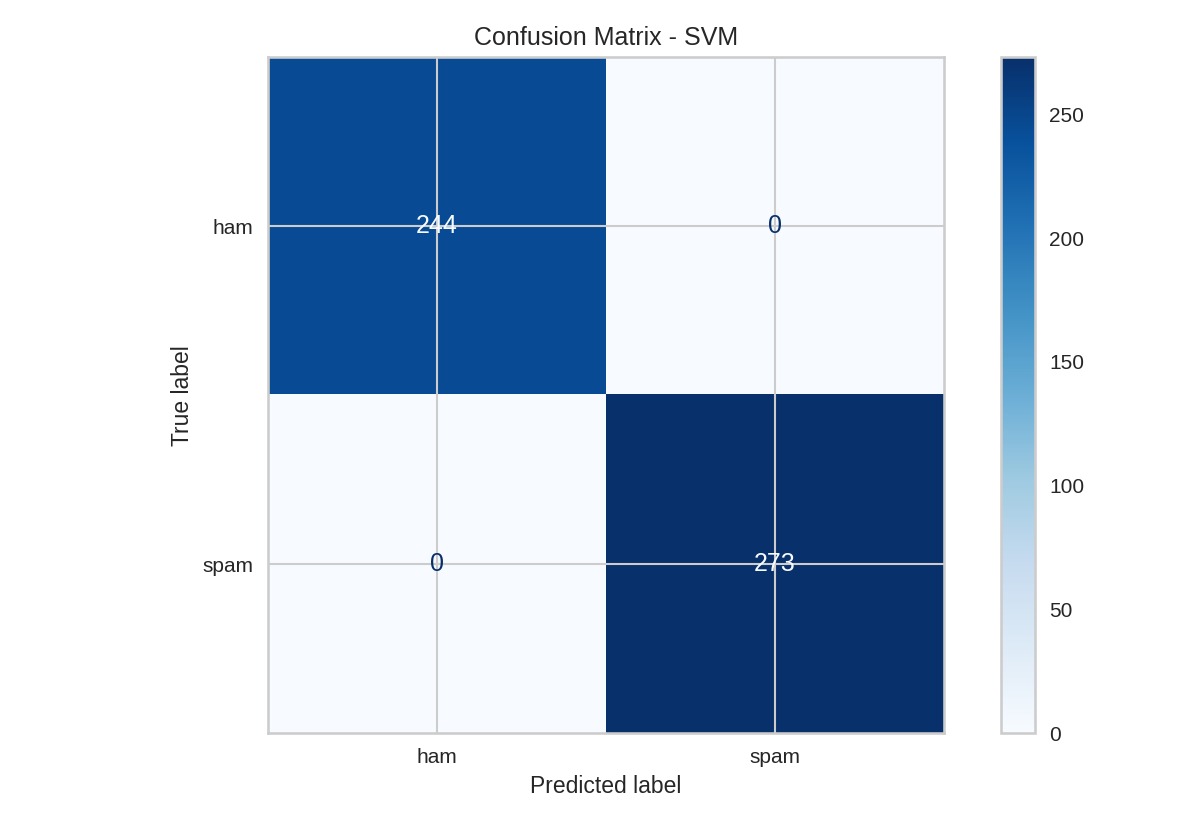}
  \caption{SVM Confusion Matrix}
  \label{fig:conf-svm}
\end{figure}

Three machine learning models (SVM, Logistic Regression, and Naive Bayes) are evaluated on the same Word2Vec features. Validation results are used for model selection, and test results for generalization assessment. These results indicate that Support Vector Machine (SVM) provides perfectly balanced and highly accurate predictions, which is consistent with its exceptional capability to handle dense Word2Vec features effectively. SVM correctly classifies all 517 of 517 test samples: 244 negatives and 273 positives, with absolutely 0 negative and 0 positive misclassifications. The lack of any errors indicates flawless and highly stable generalization performance across both classes.

Overall, SVM achieves the best machine learning performance, with an outstanding 98.74\% validation accuracy and a perfect 100\% test accuracy. Logistic Regression serves as a strong alternative, performing slightly below SVM with 97.53\% validation accuracy. Naive Bayes performs the worst, yielding 94.49\% accuracy, confirming that SVM is exceptionally well-suited for the continuous vector space generated by Word2Vec embeddings.

\subsection{LSTM Result}
The Figure 2. illustrates the training loss progression over 30 epochs, showcasing the model's convergence behavior during the training phase. Initially, the loss experiences a significant decrease, dropping sharply from approximately 0.68 to 0.20 within the first five epochs, which signifies that the model quickly adapted to the dataset's features. Although some minor fluctuations and spikes occur, most notably around epoch 22 and the overall trend remains downward. By the final stages, specifically from epoch 27 to 30, the loss stabilizes at a minimal value near 0.02. This trend indicates that the model reached a state of convergence, demonstrating high training stability and an effective error minimization process.
\begin{figure}[h]
  \centering
  \includegraphics[width=0.4\textwidth]{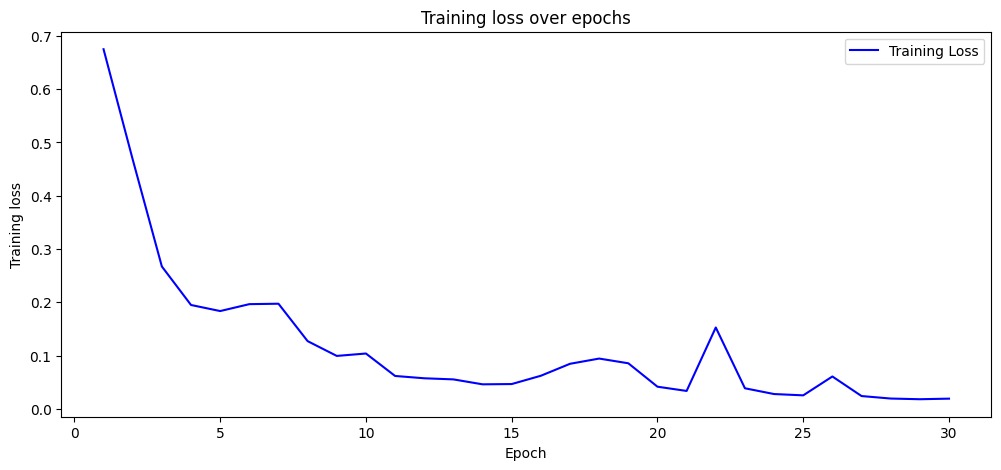}
  \caption{Training Loss}
  \label{fig:training-loss}
\end{figure}
Based on the confusion matrix presented in Figure 1, the model demonstrates high robust performance in classifying "Ham" and "Spam" messages. The evaluation results model correctly classified 234 samples as 'Ham' and 269 samples as 'Spam'. The primary diagonal of the matrix exhibits high density, indicating a strong alignment between predicted and actual labels. Specifically, the model achieved a high recall for the 'Spam' class, with only 4 instances being misclassified as 'Ham' (False Negatives). From a security perspective, this suggests that the filter is highly effective at capturing unsolicited messages. Conversely, the model recorded 10 instances where legitimate 'Ham' messages were incorrectly flagged as 'Spam' (False Positives).
\begin{figure}[h]
  \centering
  \includegraphics[width=0.5\textwidth]{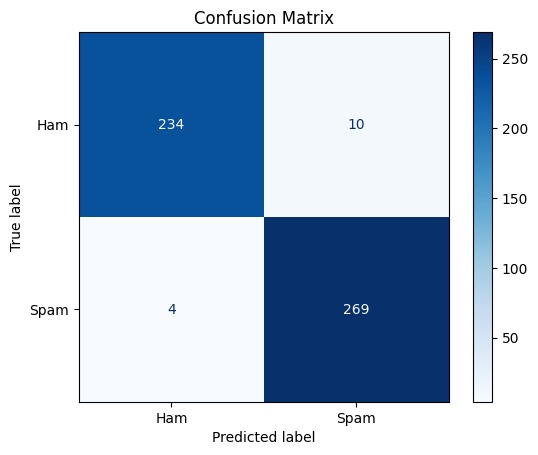}
  \caption{Confussion Matrix LSTM}
  \label{fig:conf-lstm}
\end{figure}

\begin{table}[h]
\centering
\caption{Evaluation LSTM Model}
\label{tab:classification-report}

\begin{tabular}{lcccc}
\toprule
\textbf{Class} & \textbf{Precision} & \textbf{Recall} & \textbf{F1-score} & \textbf{Support} \\
\midrule
Ham & 0.98 & 0.96 & 0.97 & 244 \\
Spam & 0.96 & 0.99 & 0.97 & 273 \\
\midrule
Accuracy &  &  & 0.97 & 517 \\
Macro Avg & 0.97 & 0.97 & 0.97 & 517 \\
Weighted Avg & 0.97 & 0.97 & 0.97 & 517 \\
\bottomrule
\end{tabular}
\end{table}
Based on performance of the Long Short-Term Memory (LSTM) model is summarized in Table. The experimental results demonstrate exceptional classification capabilities, with the model achieving an overall accuracy of 0.97. This high accuracy is consistently reflected across both classes, as evidenced by the macro and weighted average F1-scores, both standing at 0.97.

Detailed analysis reveals that the 'Ham' class achieved a precision of 0.98 and a recall of 0.96, indicating that the model is highly reliable when identifying legitimate messages. Meanwhile, the 'Spam' class exhibited a precision of 0.96 and a superior recall of 0.99. The near-perfect recall for spam detection is particularly noteworthy, as it signifies the model's ability to filter almost all unsolicited content with minimal leakage. The balanced F1-scores of 0.97 for both classes further confirm that the model does not suffer from bias toward the majority class, despite the slight difference in support (244 for Ham and 273 for Spam). These metrics, when cross-referenced with the confusion matrix in Figure 1, validate that the LSTM architecture effectively captures the long-range dependencies and semantic nuances required for accurate spam detection."

\section{Conclusion}
In this study, we have presented a comparative evaluation of four distinct algorithms, Support Vector Machine (SVM), Logistic Regression, Naive Bayes, and Long Short-Term Memory (LSTM), for email sentiment classification using Word2Vec embeddings. The experimental results demonstrate that the SVM model with a linear kernel outperforms the other approaches, achieving a superior accuracy of 98.74

While the deep learning approach using LSTM provided competitive results with 97\% accuracy and exceptional recall for spam detection, it did not surpass the efficiency and precision of the SVM in this specific vector space configuration. Our findings suggest that for tasks involving dense Word2Vec embeddings in email datasets, traditional discriminative models like SVM remain highly robust and effective

\bibliographystyle{ieeetr}
\bibliography{references}  

@article{isnansyah2024analisis,
  author    = {Gabril Isnansyah and Sutardi and Rizal Adi Saputra},
  title     = {Analisis Perbandingan Algoritma Klasifikasi Email Spam Menggunakan Long Short-Term Memory, Naïve Bayes dan Support Vector Machine},
  journal   = {ANIMATOR},
  volume    = {2},
  number    = {1},
  pages     = {1--9},
  year      = {2024},
  month     = {January--April},
  issn      = {3030-9735},
}

@article{reviantika2021analisis,
  author    = {Ferin Reviantika and Yufis Azhar and Gita Indah Marthasari},
  title     = {Analisis Klasifikasi SMS Spam Menggunakan Logistic Regression},
  journal   = {REPOSITOR},
  volume    = {3},
  number    = {4},
  pages     = {387--392},
  year      = {2021},
  month     = {August},
  issn      = {2714-7975},
  eissn     = {2716-1382},
}

@article{ainun2024klasifikasi,
  author    = {Echa Sri Ainun and Ummiyatul Inayah and Muhammad Ilmih},
  title     = {Klasifikasi Email Spam dan Ham Menggunakan Algoritma Support Vector Machine, Naive Bayes dan Logistic Regression},
  journal   = {Scientific: Journal of Computer Science and Informatics},
  volume    = {2},
  number    = {2},
  pages     = {77},
  year      = {2024},
  issn      = {3090-1391},
  doi       = {10.34304/scientific.v2.i2.399},
}

@article{wicaksana2026peningkatan,
  author    = {Hilman Singgih Wicaksana and Hargokendar Suhud},
  title     = {Peningkatan Performansi Deteksi Pesan Spam Melalui Optimasi LSTM Berbasis Word2Vec dan Grid Search},
  journal   = {JITE},
  volume    = {2},
  number    = {1},
  pages     = {21},
  year      = {2026},
  month     = {2026},
}

@article{kautsar2025analisis,
  author    = {Maugy Al Kautsar and Galet Guntoro Setiaji and Ahmad Rifa'i},
  title     = {Analisis Komparasi Kinerja LSTM dan CNN dalam Deteksi Spam Email Berbasis Deep Learning},
  journal   = {Bulletin of Computer Science Research},
  volume    = {5},
  number    = {4},
  pages     = {584--593},
  year      = {2025},
  month     = {June},
  issn      = {2774-3659},
}

@inproceedings{adduali2026komparasi,
  author    = {Muhammad Fajrin Aswad Ad-Duali and Dwika Putra Adinata},
  title     = {Komparasi Algoritma Naive Bayes dan Random Forest untuk Identifikasi Kata Berpotensi Spam},
  booktitle = {Prosiding Seminar Nasional Teknologi dan Sains Tahun 2026},
  year      = {2026},
  volume    = {5},
  address   = {Kediri, Indonesia},
  month     = {January},
  organization = {Program Studi Teknik Informatika, Universitas Nusantara PGRI Kediri},
}

@article{manguma2024analisis,
  author    = {Thiara Tri Funny Manguma and Emil Fatra},
  title     = {Analisis Performa Algoritma Klasifikasi untuk Deteksi Spam pada Email},
  journal   = {INNOVATIVE: Journal of Social Science Research},
  volume    = {4},
  number    = {3},
  pages     = {16461--16465},
  year      = {2024},
  issn      = {2807-4246},
  eissn     = {2807-4238},
}

@article{pais2022nlp,
  author    = {Sebasti{\~a}o Pais and Jo{\~a}o Cordeiro and M. Luqman Jamil},
  title     = {NLP-based Platform as a Service: A Brief Review},
  journal   = {Journal of Big Data},
  volume    = {9},
  number    = {54},
  year      = {2022},
  doi       = {10.1186/s40537-022-00603-5},
}

@article{husada2021analisis,
  title={Analisis Sentimen Pada Maskapai Penerbangan di Platform Twitter Menggunakan Algoritma Support Vector Machine (SVM)},
  author={Husada, Hendry Cipta and Paramita, Adi Suryaputra},
  journal={Teknika},
  volume={10},
  number={1},
  pages={18--26},
  year={2021}
}

@article{muttaqin2021analisis,
  title={Analisis sentimen aplikasi gojek menggunakan support vector machine dan k nearest neighbor},
  author={Muttaqin, Muhammad Nurul and Kharisudin, Iqbal},
  journal={UNNES Journal of Mathematics},
  pages={22--27},
  year={2021}
}

@article{hakim2021analisa,
  title   = {Analisa Sentimen Data Text Preprocessing pada Data Mining dengan Menggunakan Machine Learning},
  author  = {Hakim, Bhustomy},
  journal = {JBASE: Journal of Business and Audit Information Systems},
  year    = {2021},
  volume  = {4},
  number  = {2},
  pages   = {16},
}

@article{rajputnlpsentimentclinical,
  title   = {Natural Language Processing, Sentiment Analysis and Clinical Analytics},
  author  = {Rajput, Adil},
  journal = {Effat University Publication},
  year    = {n.d.},
}

@inproceedings{nakov2016semeval,
  author    = {Preslav Nakov and Alan Ritter and Sara Rosenthal and Fabrizio Sebastiani and Veselin Stoyanov},
  title     = {SemEval-2016 Task 4: Sentiment Analysis in Twitter},
  booktitle = {Proceedings of the 10th International Workshop on Semantic Evaluation (SemEval-2016)},
  pages     = {1--18},
  year      = {2016},
}

@article{elmaliyasari2025deteksi,
  title   = {Deteksi Sentimen Komentar Aplikasi Gobis Suroboyo dengan Metode Naive Bayes dan Metode Regresi Logistik},
  author  = {Elmaliyasari, Shifa and Alzam, Muhammad Arsyad and Pratiwi, Nanda Aulia and Wara, Shindi Shella May and Hindrayani, Kartika Maulida},
  journal = {JDMIS: Journal of Data Mining and Information Systems},
  volume  = {3},
  number  = {2},
  pages   = {108--116},
  year    = {2025},
}

@article{hendriyana2022analisis,
  title   = {Analisis Perbandingan Algoritma Support Vector Machine, Naive Bayes, dan Regresi Logistik untuk Memprediksi Donor Darah},
  author  = {Hendriyana and Karo Karo, Ichwanul Muslim and Dewi, Sri},
  journal = {Jurnal Teknologi Terpadu},
  volume  = {8},
  number  = {2},
  pages   = {121--126},
  year    = {2022},
}

@article{alghifari2022bilstm,
  title   = {Implementasi Bidirectional LSTM untuk Analisis Sentimen Terhadap Layanan Grab Indonesia},
  author  = {Alghifari, Dloifur Rohman and Edi, Mohammad and Firmansyah, Lutfi},
  journal = {Jurnal Manajemen Informatika (JAMIKA)},
  volume  = {12},
  number  = {2},
  pages   = {89},
  year    = {2022},
}

@article{baroto2021email,
  title   = {Email Analysis in Fraud Investigation: Digital Forensic and Network Analysis Approach},
  author  = {Baroto, Wishnu Agung},
  journal = {Asia Pacific Fraud Journal},
  volume  = {6},
  number  = {2},
  pages   = {},
  year    = {2021},
}





\end{document}